\documentclass[conference]{IEEEtran}

\usepackage[T1]{fontenc}
\usepackage{amsfonts}       % blackboard math symbols
\usepackage{nicefrac}       % compact symbols for 1/2, etc.
\usepackage{subfig}			% per usare le subfigure
\usepackage{booktabs}	% toprule midrule bottomrule
\usepackage[linesnumbered,ruled,vlined]{algorithm2e}
\usepackage{xcolor}
\usepackage[normalem]{ulem}
\usepackage{mathtools}
\usepackage[percent]{overpic} 
\usepackage{flushend} % to end last page columns with same length
\usepackage{comment}
\usepackage[noadjust]{cite}

\definecolor{black}{rgb}{0, 0, 0}
\definecolor{red}{rgb}{0.9, 0, 0}
\definecolor{green}{rgb}{0, 0.6, 0}
\definecolor{blue}{rgb}{0, 0, 0.9}
\definecolor{grey}{rgb}{0.52, 0.52, 0.51}

\SetKwInOut{KwParameter}{Parameters}
\SetKw{KwContinue}{continue}
\SetKw{KwBreak}{break}
\SetKw{KwInterval}{in}
\SetKw{KwOr}{or}
\SetKw{KwAnd}{and}
\SetKwFunction{unif}{unif}

\DeclareMathOperator*{\argmin}{argmin}

\title{\LARGE \bf
Optimizing Collaborative Robotics since Pre-Deployment via Cyber-Physical Systems' Digital Twins}
\author{Christian Cella, Marco Faroni, Andrea M. Zanchettin, Paolo Rocco%,~\IEEEmembership{Senior~Member,~IEEE}
\thanks{The authors are with Politecnico di Milano, Piazza Leonardo da Vinci, 32. Milano (Italy) {\tt\footnotesize \{marco.faroni\}@polimi.it}
}
\thanks{
This study was partially carried out within the MICS (Made in Italy – Circular and Sustainable) Extended Partnership and received funding from Next-Generation EU (Italian PNRR – M4 C2, Invest 1.3 – D.D. 1551.11-10-2022, PE00000004). CUP MICS D43C22003120001.
}
}

\begin{document}
\maketitle
\thispagestyle{empty}
\pagestyle{empty}

%%%%%%%%%%%%%%%%%%%%%%%%%%%%%%%%%%%%%%%%%%%%%%%%%%%%%%%%%%%%%%%%%%%%%%%%%%%%%%%%
\begin{abstract}

The collaboration between humans and robots requires a paradigm shift not only in robot perception, reasoning, and action, but also in the design of the robotic cell. 
This paper proposes an optimization framework for designing collaborative robotics cells using a digital twin during the pre-deployment phase. 
This approach mitigates the limitations of experience-based sub-optimal designs by means of Bayesian optimization to find the optimal layout after a certain number of iterations. 
By integrating production KPIs into a black-box optimization framework, the digital twin supports data-driven decision-making, reduces the need for costly prototypes, and ensures continuous improvement thanks to the learning nature of the algorithm. 
The paper presents a case study with preliminary results that show how this methodology can be applied to obtain safer, more efficient, and adaptable human-robot collaborative environments. 
\end{abstract}

%%%%%%%%%%%%%%%%%%%%%%%%%%%%%%%%%%%%%%%%%%%%%%%%%%%%%%%%%%%%%%%%%%%%%%%%%%%%%%%%

\begin{comment}
This parapgraph is commented out.
This parapgraph is commented out.
\end{comment}

% This line is commented out.

\section{Introduction}

%Collaboration between humans and robots requires a paradigm change in robots' perception, reasoning, and action. 
%This shift encompasses different levels of abstraction and reactivity. 
%For example, the system should find a suitable sequence of operations (task planning), assign them to the agents (task assignment), and schedule their execution (scheduling). 
%At run-time, the execution of the operations should be adapted to the human and robot's state (motion planning and replanning). 
%All these steps should also consider the variability of humans and the effect of simultaneous operations carried out by the robots and the workers.
%
%Existing works usually aim to improve the efficiency of an existing collaborative cell.
%For example, many focused on task planning \cite{Makris:task-planning-hrc,Toussaint:ergonomic-planning}, scheduling \cite{Makris:scheduling-hrc,Zanchettin:scheduling}, motion planning \cite{Lasota:hamp,Berenson:human-robot-planning}, and control \cite{haw-control,Lippi:T-CST}. 
%The goal is to reduce cycle time or to improve the human experience by including human factors in the robot decision-making.
%However, these approaches are limited by the sub-optimal design of the cell, which was not designed with such methodologies in mind.
%This is why we claim that the optimization of the collaborative process should begin during the design phase.

The collaboration between humans and robots necessitates a paradigm shift in the way robots perceive, reason, and act. This shift must address multiple levels of abstraction and reactivity. For instance, the system must identify a suitable sequence of operations (task planning), assign these tasks to the appropriate agents (task assignment), and schedule their execution (scheduling). 
Additionally, during run-time, the execution of these operations must be adapted based on the states of both humans and robots (motion planning and possibly replanning) \cite{riedelbauch2023benchmarking}. 
%These processes must also consider human variability and the effects of simultaneous operations carried out by both robots and workers .

The need for optimized methods in designing robotics cells for human-robot collaboration is of paramount importance. 
Existing research typically focuses on enhancing the efficiency of pre-existing collaborative cells. For example, several studies focused on task planning \cite{Makris:task-planning-hrc,Toussaint:ergonomic-planning, Sandrini_ETFA2022}, scheduling \cite{Makris:scheduling-hrc,Zanchettin:scheduling}, task assignment \cite{fusaro2021human,kiyokawa2023difficulty,yao2024task}, motion planning \cite{Lasota:hamp,Berenson:human-robot-planning, Faroni_hamp_RAL2022, Tonola_ROMAN2021}, and control \cite{haw-control,Lippi:T-CST, Faroni_ETFA2019}. 
These efforts aim to reduce cycle times and improve human experience by incorporating human factors into robot decision-making. However, these approaches often fall short due to the sub-optimal initial design of the cell, which was not created with these advanced methodologies in mind, but it just represents a reasonable trade-off based on the experience of the designer.

We argue that the optimization of the collaborative process must begin during the design phase. 
By integrating optimization techniques early on, it is possible to create collaborative cells that are inherently more efficient and better suited for human-robot interaction. 
This proactive approach ensures that the design of the cell itself supports optimal performance, rather than trying to retrofit improvements onto a fundamentally sub-optimal setup \cite{Cella_MetroXraine}.

This paper deals with the optimization of the layout of the production cell, including the human and robot positioning, the number and localization of tools and pieces, and the spatial layout of the station. 
The technological solution proposed herein consists of a flexible digital twin of the collaborative process. 
The digital twin allows for fast re-configuration and evaluation of different layouts in terms of production KPIs (e.g., takt time, cycle time) and human-centric KPIs (e.g. ergonomics scores). 
To this end, we propose a black-box optimization framework where the black-box cost function is represented by the selected KPIs and is evaluated by running high-fidelity process simulations within a Bayesian optimization framework. 

Using a digital twin to optimize the robotics cell before deployment offers several significant advantages. One major benefit is the enhanced simulation of human movements and ergonomics. A digital twin can accurately replicate human actions, ensuring that the cell accommodates human factors effectively, leading to a safer and more comfortable working environment for operators. It also allows for the prediction and mitigation of human-robot interaction issues, such as potential collisions and awkward postures, by identifying these problems in the virtual environment before the physical cell is constructed.

%Furthermore, a digital twin facilitates the optimization of task assignments and scheduling by running numerous simulations that account for both robot capabilities and human constraints. This ensures a balanced workload and optimal productivity. The ability to simulate various scenarios and conditions enhances the adaptability and flexibility of the cell design, making it capable of adjusting to changes in tasks or workforce dynamics.

Additionally, identifying and resolving design issues in the virtual environment reduces the need for costly physical prototypes and iterative modifications, thus accelerating the deployment process and lowering overall costs. The integration of human factors into the simulation ensures that the design is human-centric, leading to improved operator satisfaction, reduced fatigue, and a lower risk of work-related injuries.

The virtual replica of the system also supports data-driven decision-making by collecting and analyzing data from simulations, providing valuable insights that inform the design process. By proactively identifying potential bottlenecks and inefficiencies, designers can address these issues early, leading to smoother and more efficient operations. 
%The digital twin can also be used for training human operators in a virtual environment, ensuring they are familiar with the cell and can integrate seamlessly with robotic systems from the outset.

Finally, the digital twin paradigm enables continuous improvement by being updated and refined based on real-world data and feedback, ensuring that the robotics cell remains efficient and effective over time. %In summary, leveraging a digital twin for optimizing the design of a robotics cell before deployment enhances the efficiency and safety of human-robot interactions and ensures a cost-effective, adaptable solution that meets both human and operational needs.

The main contribution of this paper is a new Bayesian optimization framework for optimal layout design of collaborative cells.
The framework is based on the coupling of a high-fidelity simulation environment that virtualizes the most important aspects of the collaborative process and state-of-the-art Bayesian optimizers, that represent promising constructs when dealing with the optimization of functions that can either be unknown or difficult to evaluate.
We discuss preliminary results on a manufacturing-oriented case study with shared workspace between a lightweight robot (Universal Robots UR5e) and an operator in a real-world cell.

The paper is structured as follows. 
Section \ref{sec:related-works} analyzes the state of the art of collaborative cells design. 
Section \ref{sec:method} describes the proposed approach.
Section \ref{sec:case-study} analyzes a manufacturing case study involving a collaborative robot and a human operator performing collaborative tasks.
Finally, Section \ref{sec:conclusions} draws conclusions and discusses future works.

\section{Related Works}\label{sec:related-works}

Design of robotic cells is a long studied problem driven by the heavy impact of sub-optimal layout on production throughput \cite{dawande2007throughput}.
Most approaches focused on mass-production lines (e.g., automotive industries), where the cell layout involves choosing the mounting positions of multiple robots to minimize the average steady-state cycle time in repetitive production \cite{kamoun1999scheduling}.
Other approaches play with task positioning to improve the quality of the technological process (e.g., laser cutting) by maximizing the robot dexterity \cite{mutti2021towards}.
These scenarios are characterized by the non-collaborative robots performing repetitive actions in a static and structured environment.

Collaborative robotics poses new challenges in the design of the cell layout because of the presence of human operators.
In most practical scenarios, the design of a collaborative cell is evaluated manually by an operator \cite{malik2021digital}. 
Various layouts are simulated on the digital twin, with changes made to the type of robot, fixtures, and their relative positions each time. 
This heuristic method depends solely on the layout designer's experience and often lead to sub-optimal solutions. To our knowledge, there are no existing studies that specifically address a digital-twin-guided procedure for determining the optimal layout of a collaborative working cell without relying on real-world data. 
%Consequently, an iterative procedure must be established during the pre-deployment phase, leveraging a specific optimization algorithm while simultaneously considering task scheduling and resource allocation among the different agents.

Tay and Ngoi \cite{tay1996optimising} sought to identify the optimal layout of a robotic work cell using an offline matrix approach to represent dimensions and scheduling activities. 
On the other hand, Tsarouchi et al. \cite{tsarouchi2016decision} developed a decision-making framework for layout generation based on digital twin simulations, though human intervention is still needed to manually adjust simulation parameters. 
Another research line focuses on the calibration of virtual models to fine-tune the digital twin's behavior. 
Unlike previous implementations, these approaches require an existing physical unit and sensor data collected from the area. 
After defining a cost functional that incorporates both real and simulated data, it is minimized using a combination of heuristic and statistical algorithms.
The pipeline introduced by Chakrabarty, Wichern, and Laughman \cite{chakrabarty2021attentive} serves as an example of an offline model calibration procedure based on a modified Bayesian optimization scheme. 
In this scheme, the kernel function is not approximated by a surrogate as standard settings would dictate; instead, an Attentive Neural Process is applied following the selection of the data batch. 
Another example of model calibration within a virtual environment is the work of Chakrabarty, Bortoff, and Laughman \cite{chakrabarty2022simulation}, where the Bayesian algorithm is guided by failure regions to avoid while optimizing model parameters.

With respect to the works discussed above, our methodology is not used for the calibration of the unknown parameters of a given model, nor does it consider human intervention at any stage of the process. 
Instead, the main goal of this work is to create an optimization framework that runs offline and can generate an optimized collaborative cell before the actual deployment.
The \textit{black-box} nature of the Bayesian optimizer allows to drive the search of a new candidate solution to the most promising regions of the domain thanks to the learning-based feedback resulting from each simulation.
The result is a ready-to-deploy cell layout designed for cycle-time (and possibly other objectives) minimization.

\begin{figure*}[tpb]
	\centering
	\includegraphics[trim = 0cm 0cm 0cm 0cm, clip, angle=0, width=0.8\textwidth]{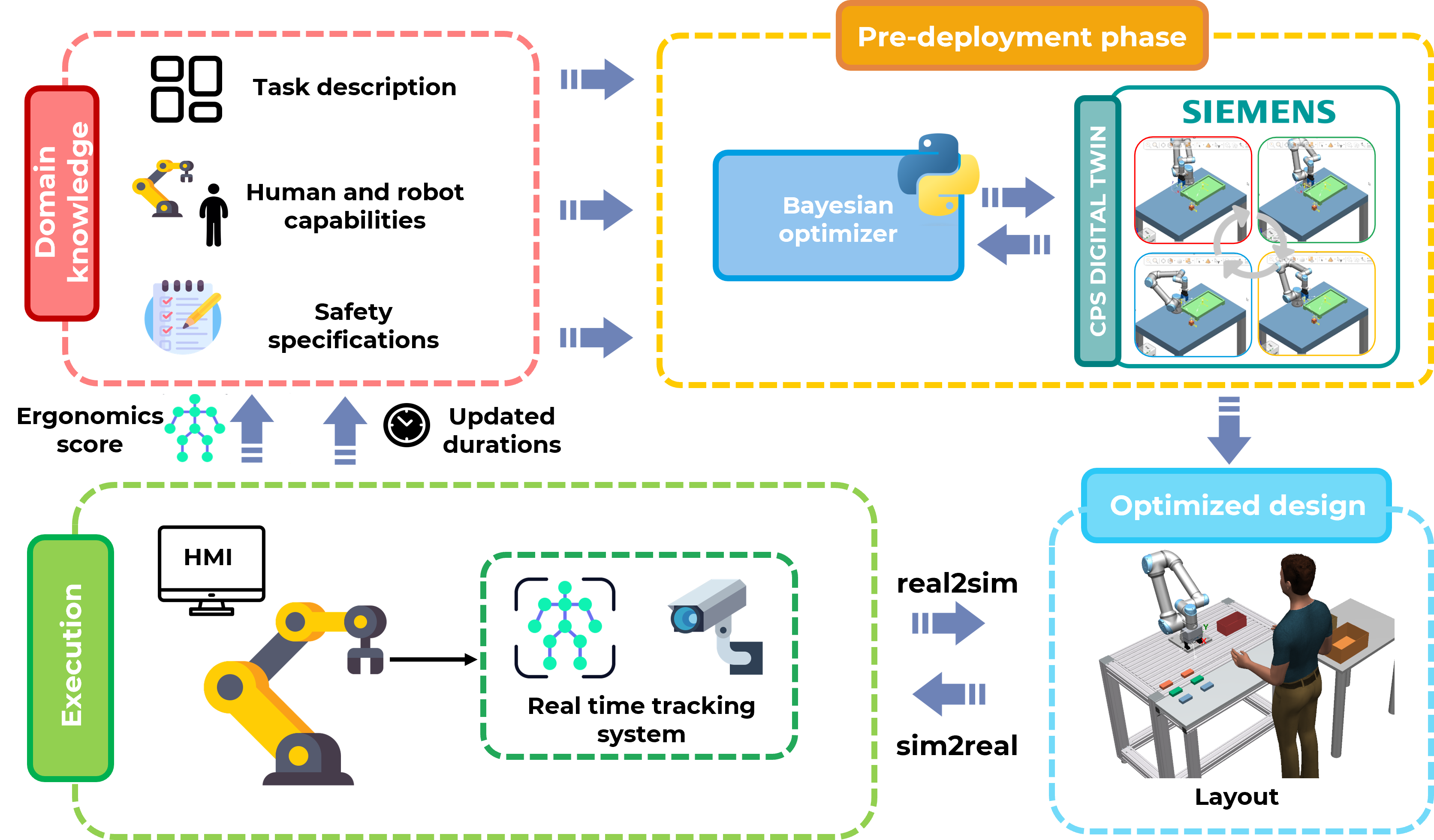}
	\caption{Overall framework presented: the digital twin of the collaborative cell is used both at the level of $\textit{pre-deployment}$ as a simulator and as the actual digital mock-up of the system during the $\textit{executional}$ stage.}
	\label{fig:framework}
\end{figure*}

\section{Approach} \label{sec:method}

\begin{figure*}[tpb]
	\centering
	\includegraphics[trim = 0cm 0cm 0cm 0cm, clip, angle=0, width=0.8\textwidth]{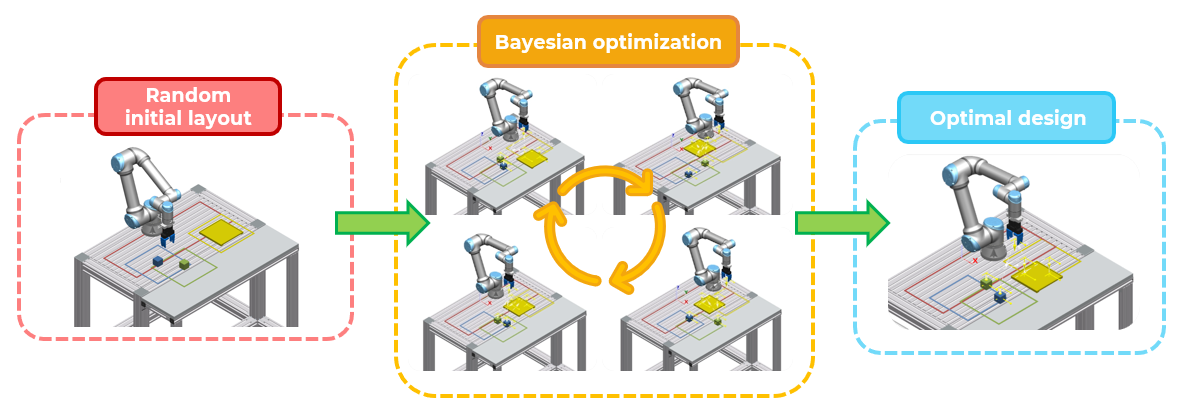}
	\caption{Illustrative scheme of the Bayesian optimization algorithm: an initial layout is selected and, with the aim of minimizing a specific KPI, a final solution is proposed by the algorithm.}
	\label{fig:optimizer}
\end{figure*}

Our approach intertwines a high-fidelity human-robot simulator, Tecnomatix Process Simulate 2307 by Siemens \cite{process_simulate}, with a Bayesian optimization framework to optimize the design of collaborative robotics cells. 
This section details the integration process, the choice of the simulation setup and the optimization procedure.
The overall framework of our methodology is illustrated in Figure \ref{fig:framework}. 
This work focuses on the integration of the Bayesian Optimization module to generate the optimized design, starting from domain knowledge about the process.

\subsection{Integration of Simulator and Optimizer}

\begin{figure}[tpb]
	\centering
	\includegraphics[trim = 0cm 0cm 0cm 0cm, clip, angle=0, width=0.8\columnwidth]{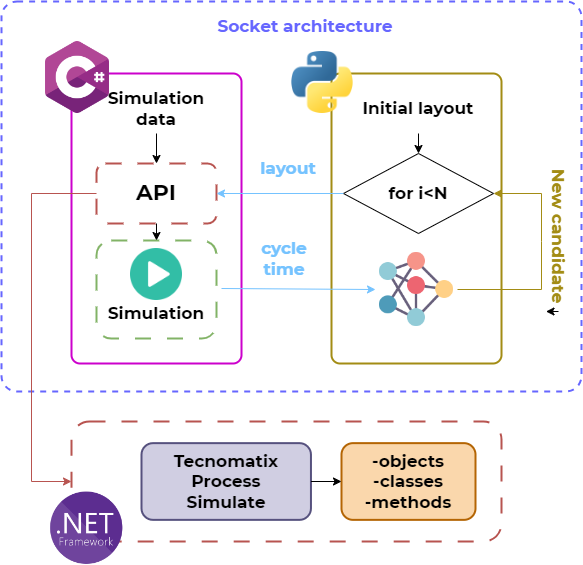}
	\caption{Scheme depicting the socket communication.}
	\label{fig: socket}
\end{figure}

To effectively combine the capabilities of the Tecnomatix Process Simulate with a Python-based Bayesian optimizer, we developed custom APIs. These codes serve as a bridge, allowing seamless communication between the C\#-based commercial simulator and state-of-the-art Bayesian optimization frameworks such as \texttt{PyBO} \cite{balandat2020botorch} and \texttt{scikit} \cite{scikit-learn}.
This integration enables to compute the cycle time associated to each new simulation and return it to the Bayesian optimizer, in order to inform subsequent iterations. This procedure is repeated until the maximum number of iterations N is reached. The overall communication structure is encapsulated in a more complex software construct, namely a socket, that involves the data exchange according to the TCP/IP protocol following the scheme represented in Figure \ref{fig: socket}.

\subsection{Simulation Setup and Optimization Objectives}

In Tecnomatix Process Simulate, we model the collaborative robotics cell, incorporating detailed representations of both human and robot movements. 
For the latter, we devised custom functionalities to represent human actions in a more realistic way. 
In this way, the simulator is capable of high-fidelity simulations that account for various task sequences, agent assignments, and real-time state adaptations, including human variability and the impact of concurrent operations. 
For the purpose of this work, we assume the task assignment is given.
Considering task scheduling and allocation, as well as the possibility of performing tool changes at run-time, is left as a possible future work.
%as shown in the images taken from the real world in Figure \ref{fig: implementation}, the manipulator is limited to grasping the items on the table with an electric gripper. 
The primary objective of our optimization is to minimize cycle time, thereby enhancing overall efficiency. 
Additionally, we incorporate human-centered factors into the optimization criteria, such as ergonomics and perceived safety, to ensure a human-centric design. 
These factors are quantified using specific metrics within the simulator, allowing for a comprehensive evaluation of each design iteration and are enforced as additional constraints on the acquisition function.

\subsection{Bayesian Optimization Process}

Bayesian optimization is a very well known technique \cite{frazier2018tutorial} which is utilized to navigate a complex design space effectively. In case of a single objective constrained optimization problem the formulation is as follows:

% Equation of the bayesian optimization
\begin{equation}
\begin{aligned} \textbf{x}_{opt} =\argmin_{\textbf{x}\in\mathcal{X}\subseteq\mathbb{R}^{\text{D}}} \quad & f(\textbf{x})\\
\textrm{subject to} \quad & \textbf{g}(\textbf{x})\leq\textbf{0}
\end{aligned}
\label{equation:bayesian_optimization}
\end{equation}
In our work, the design $\textbf{x}\in\mathbb{R}^{\text{D}}$ is a vector composed of the x and y coordinates of the instances present in the scene, namely the robot and the objects to be picked. Moreover, the search space $\mathcal{X}$ for the x and y coordinates is bounded inside rectangular areas, of the type presented in Figure \ref{fig:optimizer}. In order to avoid that the new layout suggested by the algorithm results in objects that overlap on each other, twelve inequality constraints $\textbf{g}(\textbf{x})$ are introduced in the process, in the form of tolerable Euclidean distances between the items in the scene:

% Equation of the Euclideaan distances
\begin{equation}
\begin{aligned}
\resizebox{0.44\textwidth}{!}{$|\sqrt{(x_i-x_j)^2 + (y_i-y_j)^2} - \frac{d_{max_{i,j}} + d_{min_{i,j}}}{2}|\geq\frac{d_{max_{i,j}} - d_{min_{i,j}}}{2}$}
\end{aligned}
\label{equation:euclidean_distances}
\end{equation}\

In equation \ref{equation:euclidean_distances}, subscripts i and j represent couples of instances in the simulation environment ($\textit{i.e.}$ the robot and one of the items to be picked), while $d_{max_{i,j}}$ and $d_{min_{i,j}}$ are respectively the maximum and minimum distances between the i-th and the j-th resource.

The iterative nature of the process allows to substitute the evaluation of the unknown black-box function $f$ (with reference to equation \ref{equation:bayesian_optimization}) with the optimization of a suitable acquisition function $\hat{f}_{acq}$. In this case, the optimal value of the design $\textbf{x}$ is replaced by the new tentative candidate $\textbf{x}_{next}$:

% Equation of iterative procedure
\begin{equation}
\begin{aligned} \textbf{x}_{next} =\argmin_{\textbf{x}\in\mathcal{X}\subseteq\mathbb{R}^{\text{D}}} \quad & \hat{f}_{acq}(\textbf{x})\\
\textrm{subject to} \quad & \textbf{g}(\textbf{x})\leq\textbf{0}
\end{aligned}
\label{equation:bayesian_iterative}
\end{equation}
The choice of the acquisition function depends on the problem under exam: for the purpose of this work, the function \textit{Upper Confidence Bound} (UCB) was selected:

% equation of the acquisition function
\begin{equation}
\begin{aligned}
\hat{f}_{acq} = \mu(\textbf{x}) + \kappa\cdot\sigma(\textbf{x})
\end{aligned}
\label{equation:acquisition}
\end{equation}
In the equation above, $\mu$ and $\sigma$ are the mean value and the standard deviation associated to the augmented vector of observations, while $\kappa$ is a sigmoidal function capable of reducing the exploration of the algorithm after a certain number of iterations:

% equation of the sigmoidal function
\begin{equation}
\begin{aligned}
\kappa = \frac{\kappa_0}{1+e^{-a \cdot{(b - k)}}}
\end{aligned}
\label{equation:sigmoid}
\end{equation}
For the problem under exam, the initial value of the sigmoid $\kappa_0$ was selected equal to 2, while the two remaining hyperparameters, namely the scaling factor $a$ and the translation factor $b$, were set to 0.1 and to the 75$\%$ of the maximum number of iterations ($N_{sim}$).

The process begins with an initial set of design configurations, which are simulated in Tecnomatix Process Simulate. Once a simulation is completed, the cycle time is computed and returned to the Bayesian optimizer, that will augment the set of observations to fit a new Gaussian process. Based on this refined set of observations it is possible to infer the new mean value and standard deviation to be inserted in equation \ref{equation:acquisition}, and the new layout $\textbf{x}_{next}$ can be determined.

This iterative process continues until specific convergence criteria are met or a predefined number of iterations is reached.
The overall workflow can be summarized as follows:
\begin{enumerate}
    \item \textbf{Initialization}: Start with an initial design set, selected based on prior knowledge or random sampling.
    \item \textbf{Simulation}: Run the initial set of designs in Tecnomatix Process Simulate, capturing relevant metrics.
    \item \textbf{Optimization Loop}:
    \begin{itemize}
        \item Return simulation results to the Bayesian optimizer.
        \item Update the probabilistic model with new data.
        \item Select the next design configurations based on the selected acquisition function.
        \item Run simulations for the new configurations.
    \end{itemize}
    \item \textbf{Convergence}: Repeat the optimization loop until the desired performance is achieved or the iteration limit is reached.
\end{enumerate}
Figure \ref{fig:optimizer} shows a simplified layout that is subsequently elaborated by the Bayesian optimizer. The main input of the iterative optimization is the initial random layout that is selected based on the operator's experience, together with the domain knowledge that fully describes the preliminary information about a collaborative process (red block in Figure \ref{fig:framework}). After a given number of iterations the final solution is outputted by the framework, and represents the bridge between the pre-deployment phase and the real-time execution of the cell.

By integrating high-fidelity simulations with Bayesian optimization, our methodology allows for efficient and effective optimization of collaborative robotics cells, ensuring both operational efficiency and human-centric design considerations.

\section{Case study} \label{sec:case-study}

We conducted preliminary tests on a case study utilizing a collaborative robot with a human operator to execute pick and packaging operations. 
This case study serves as an illustrative example of how our framework can be applied in real-world scenarios. 

%By integrating the UR5e robot and the human operator within our optimization framework, we aim to optimize the layout and operation of the collaborative robotics cell to achieve enhanced efficiency.

\begin{figure*}[tpb]
	\centering
    \includegraphics[trim = 0.2cm 0.2cm 0.2cm 0.2cm, clip, angle=0, width=0.24\textwidth]{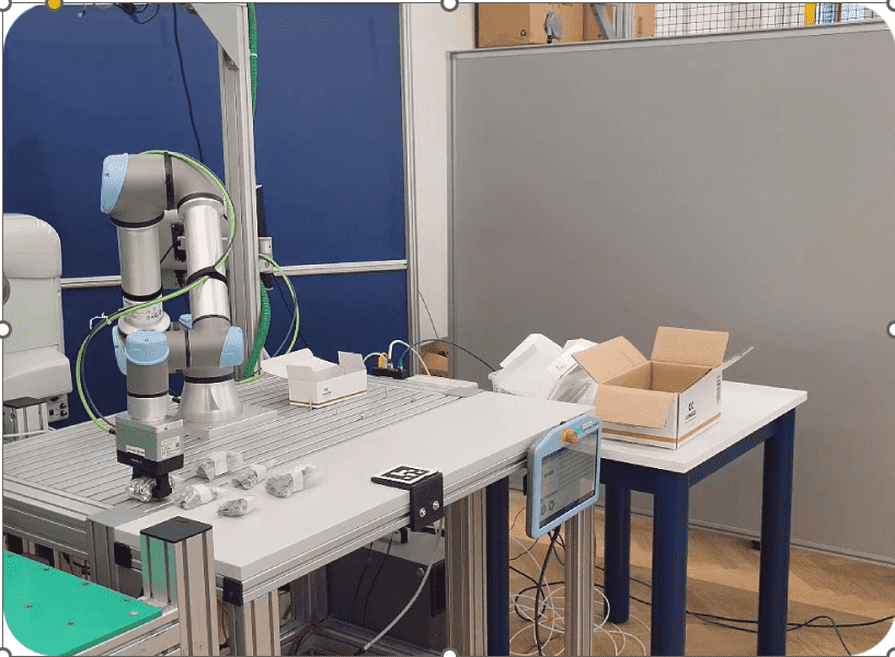}\,
 	\includegraphics[trim = 0.2cm 0.2cm 0.2cm 0.2cm, clip, angle=0, width=0.24\textwidth]{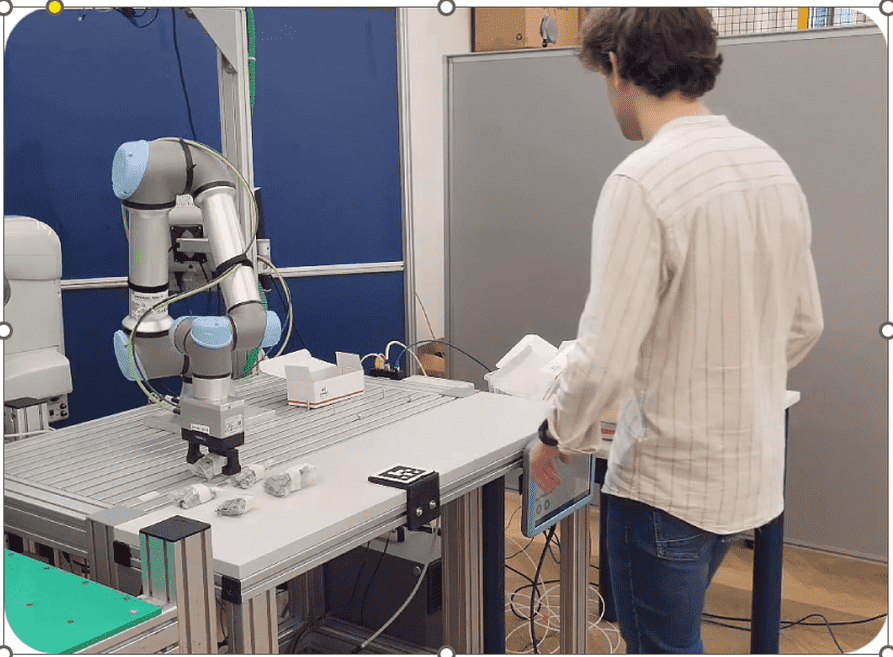}\,
  	\includegraphics[trim = 0.2cm 0.2cm 0.2cm 0.2cm, clip, angle=0, width=0.24\textwidth]{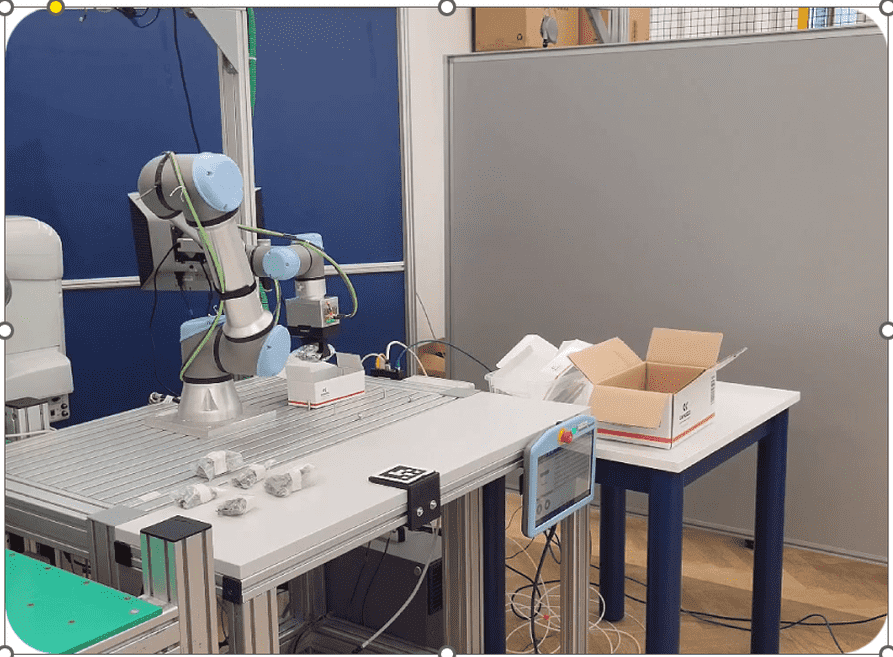}\,
   	\includegraphics[trim = 0.2cm 0.2cm 0.2cm 0.2cm, clip, angle=0, width=0.24\textwidth]{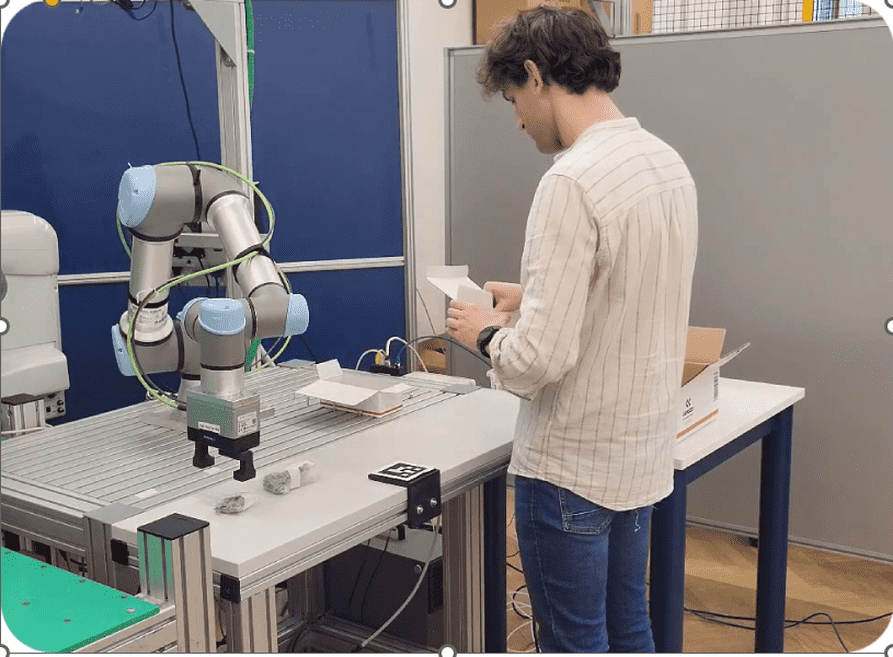}
    \\
    \vspace{0.2cm}
    \includegraphics[trim = 0.2cm 0.2cm 0.2cm 0.2cm, clip, angle=0, width=0.24\textwidth]{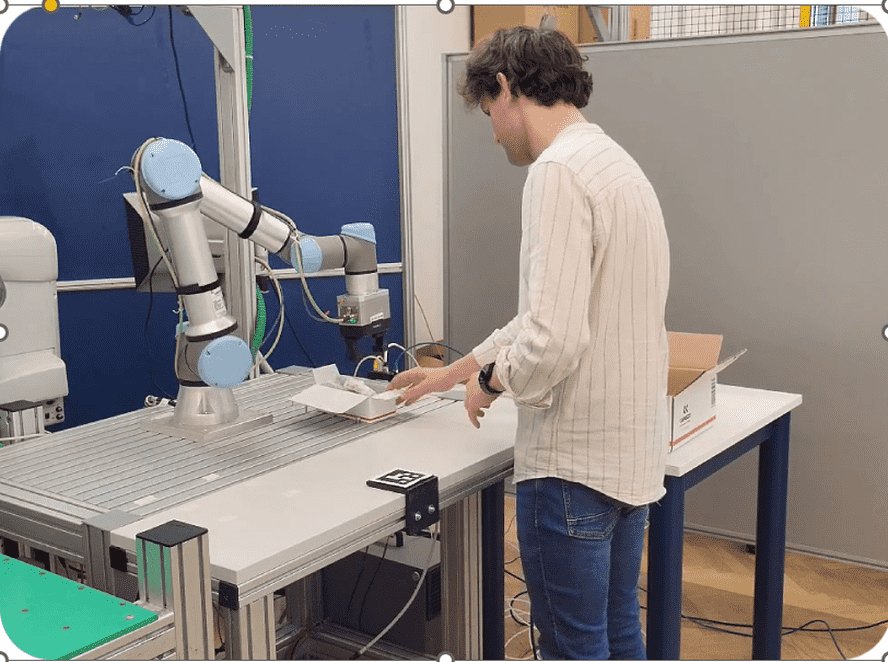}\,
 	\includegraphics[trim = 0.2cm 0.2cm 0.2cm 0.2cm, clip, angle=0, width=0.24\textwidth]{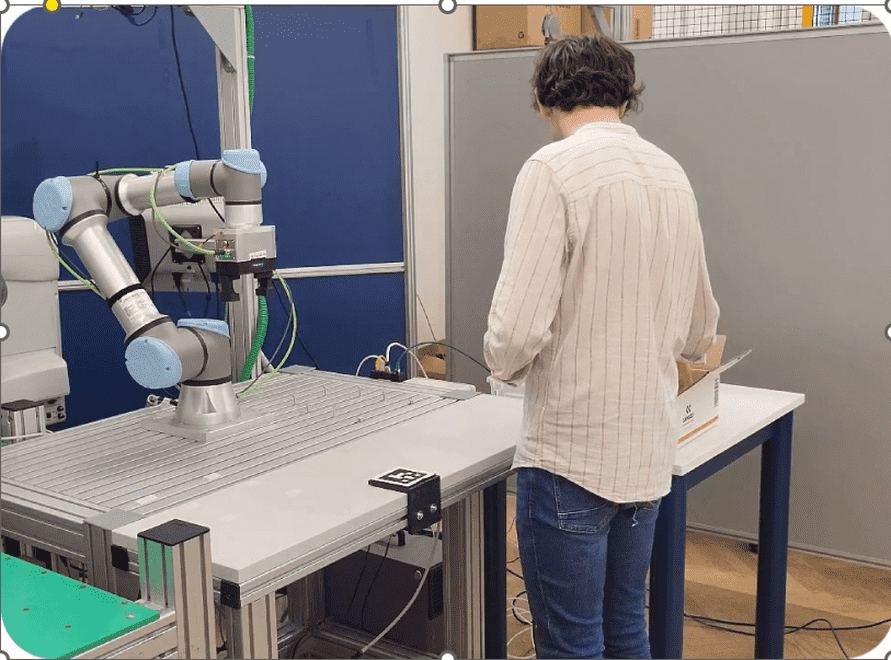}\,
  	\includegraphics[trim = 0.2cm 0.2cm 0.2cm 0.2cm, clip, angle=0, width=0.24\textwidth]{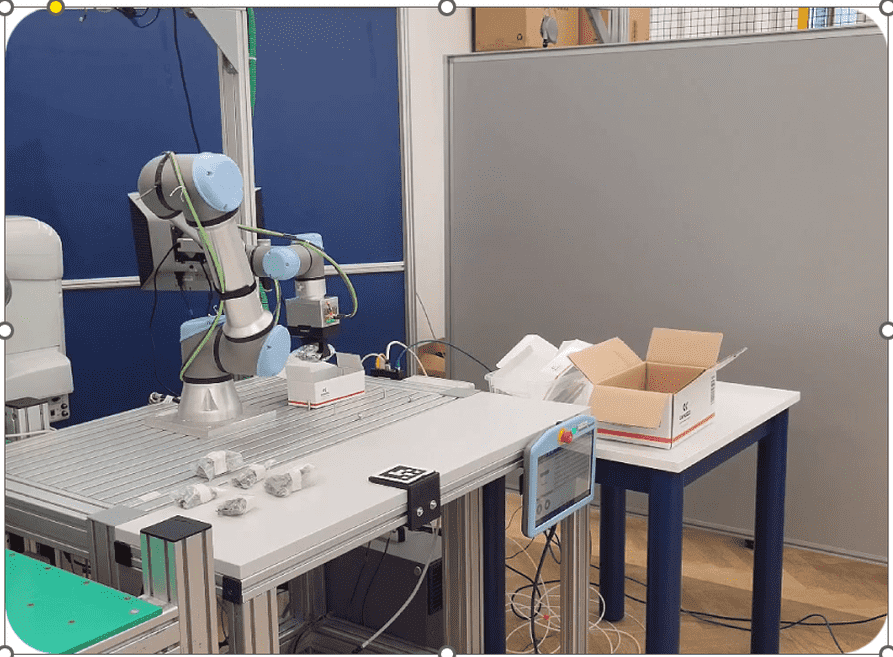}\,
   	\includegraphics[trim = 0.2cm 0.2cm 0.2cm 0.2cm, clip, angle=0, width=0.24\textwidth]{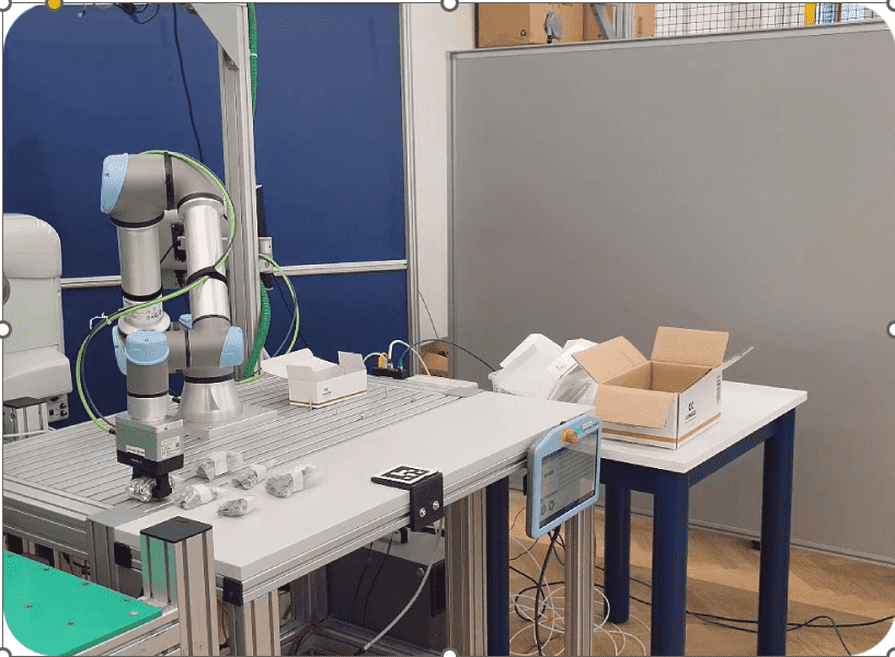}
	\caption{An example of executed process with the optimized layout. }
	\label{fig: implementation}
\end{figure*}

Figure \ref{fig: implementation} shows the collaborative cell and the tasks.
The cell is composed of a collaborative robot, model Universal Robots UR5e, an electro-mechanical gripper, model Schunk EGK-40, and an RGB-D camera, model Realsense D435i, for real-time tracking of the human motion.
An external PC communicates with the robotic system via the ROS1-Noetic middleware \cite{quigley2009ros}.
Note that we use ROS to trigger the robot actions, yet the execution of each action is performed by the robot controller (i.e., ROS sends \texttt{ur\_script} files to the robot) to leverage Process Simulate's Realistic Robot Simulation (RCS).
RCS is is a standard that allows providing accurate robot simulations and cycle time estimates.
It allows integrating original controller software into simulations and offline programming software: in this way, we ensure a high-fidelity matching between the robot movements and its digital twin.

The process consists of several pick and packaging operations: the robot shall pick six objects from the the table and place them into three containers.
In the meantime, the operator first places the containers on the table and, once a container is full, moves them to an outbound location.
Snapshots of the collaborative process are shown in Figure \ref{fig: implementation}.

In this case study, the Tecnomatix Process Simulate simulator is utilized to model the pick and packaging operations, taking into account the capabilities of the UR5e robot and the interactions between the robot and the human operator. 
The simulator allows for the simulation of various task sequences, agent assignments, and real-time state adaptations, enabling a comprehensive evaluation of the collaborative process.

The Bayesian optimization framework is then employed to navigate the complex design space and identify optimal configurations for the collaborative robotics cell. 
The optimization variables are the robot mounting position on the table, the position of the objects to be picked, and the position of the placing boxes.
In this test, we focus on process time efficiency, yet other relevant metrics such as ergonomics and perceived safety will be object of future works.

\begin{figure}[tpb]
	\centering
	\includegraphics[trim = 0cm 0cm 0cm 0cm, clip, angle=0, width=0.8\columnwidth]{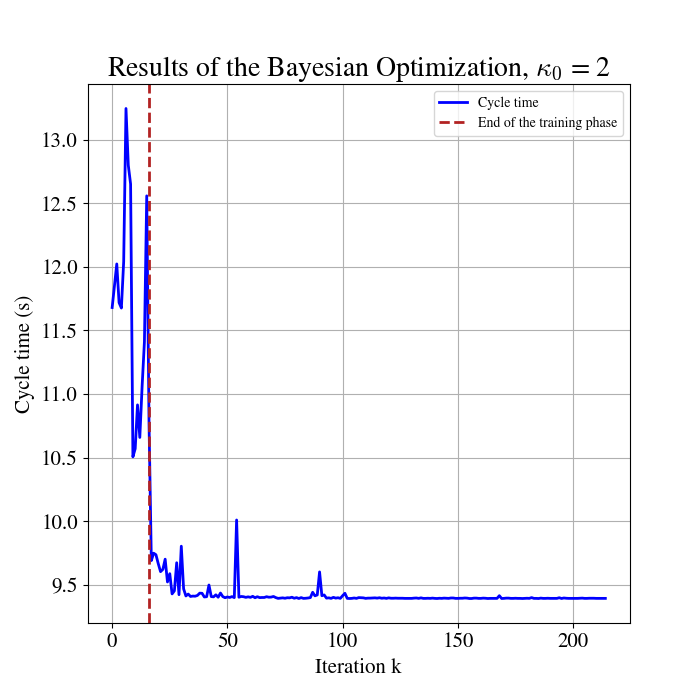}
	\caption{Cycle time trend over iterations  of the Bayesian optimizer.}
	\label{fig: cycle-time}
\end{figure}

Figure \ref{fig: cycle-time} shows the trend of the cost function over iterations.
The first 20 iterations serve to initialize the surrogate model (i.e., a Gaussian Process) with random samples.
After that, the optimizer comes into play and chooses the next sample according to the acquisition function as described in Section \ref{sec:method}.
The Bayesian optimizer takes about 100 iterations to converge to a minimum, i.e., to find the optimized cell layoutl, and this represents almost half of the maximum number of iterations imposed at the beginning ($N_{sim}$ = 200).
The optimized layout found by our approach leads to an expected process cycle time of about 9.4 seconds thanks to the relocation of the robot and the pieces.
This represents an improvement of about -18\% with respect to the average cycle time obtained with random layouts.
Figure \ref{fig: implementation} shows an execution of the process with the optimized layout in the real-world cell.
%A video is also available at XXXX TODO.

\section{Conclusions}\label{sec:conclusions}

This paper presented a novel approach for the optimization of collaborative robotics cells by utilizing a digital twin during the pre-deployment phase. 
This method addresses the critical need for improved design processes that accommodate the complexities of human-robot collaboration. 
By integrating advanced simulations of human movements and ergonomics, the digital twin allows for realistic human-robot interaction and estimation of the cycle time.

The proposed black-box optimization framework can accommodate for production and human-centric KPIs to evaluate various cell layouts. 
This enables a comprehensive assessment that balances efficiency, productivity, and operator well-being. 

The benefits of this approach are manifold. It reduces the need for physical prototypes and iterative modifications, thereby lowering costs and accelerating deployment.

The case study demonstrated the practical application of this methodology, showcasing its potential to transform the design and deployment of collaborative robotics cells. 
Future work will focus on refining the optimization algorithms, multi-objective cost function (involving ergonomics and safety) and providing real-world comparison with state-of-the-art approaches.
The final goal of this framework is to foster a human-centric design philosophy, which enhances operator satisfaction, reduces fatigue, and minimizes the risk of work-related injuries. 
The data-driven decision-making facilitated by the digital twin will ensure continuous improvement and long-term efficiency of the robotics cell.

\section*{Acknowledgement}

This study was carried out within the MICS (Made in Italy – Circular and Sustainable) Extended Partnership and received funding from Next-Generation EU (Italian PNRR – M4 C2, Invest 1.3 – D.D. 1551.11-10-2022, PE00000004). CUP MICS D43C22003120001

\bibliographystyle{IEEEtran}
\bibliography{bibliography}

\end{document}